\documentclass[journal]{IEEEtran}

\usepackage{times}
\usepackage{epsfig}
\usepackage{graphicx}
\usepackage{amsmath}
\usepackage{amssymb}
\usepackage{makecell}
\usepackage{todonotes}
\usepackage{tablefootnote}
\usepackage{longtable}
\usepackage{caption}
\usepackage{enumitem}
\usepackage[symbol]{footmisc}
\usepackage{comment}
\usepackage{adjustbox}
\usepackage[switch,columnwise]{lineno}
\usepackage[normalem]{ulem}
\RequirePackage{color}\definecolor{RED}{rgb}{1,0,0}\definecolor{BLUE}{rgb}{0,0,1} 
\usepackage{multirow}

\newcommand{\RNum}[1]{\uppercase\expandafter{\romannumeral #1\relax}}
\newcommand{\etal}{\emph{et al.}}

\RequirePackage[normalem]{ulem} 
\RequirePackage{color}\definecolor{RED}{rgb}{1,0,0}\definecolor{BLUE}{rgb}{0,0,1} 
\providecommand{\DIFaddbegin}{} 
\providecommand{\DIFaddend}{} 
\providecommand{\DIFdelbegin}{} 
\providecommand{\DIFdelend}{} 
\providecommand{\DIFaddbeginFL}{} 
\providecommand{\DIFaddendFL}{} 
\providecommand{\DIFdelbeginFL}{} 
\providecommand{\DIFdelendFL}{} 
\newcommand{\DIFscaledelfig}{0.5}
\RequirePackage{settobox} 
\RequirePackage{letltxmacro} 
\newsavebox{\DIFdelgraphicsbox} 
\newlength{\DIFdelgraphicswidth} 
\newlength{\DIFdelgraphicsheight} 
\LetLtxMacro{\DIFOincludegraphics}{\includegraphics} 
\newcommand{\DIFaddincludegraphics}[2][]{{\color{blue}\fbox{\DIFOincludegraphics[#1]{#2}}}} 
\newcommand{\DIFdelincludegraphics}[2][]{
	\sbox{\DIFdelgraphicsbox}{\DIFOincludegraphics[#1]{#2}}
	\settoboxwidth{\DIFdelgraphicswidth}{\DIFdelgraphicsbox} 
	\settoboxtotalheight{\DIFdelgraphicsheight}{\DIFdelgraphicsbox} 
	\scalebox{\DIFscaledelfig}{
		\parbox[b]{\DIFdelgraphicswidth}{\usebox{\DIFdelgraphicsbox}\\[-\baselineskip] \rule{\DIFdelgraphicswidth}{0em}}\llap{\resizebox{\DIFdelgraphicswidth}{\DIFdelgraphicsheight}{
				\setlength{\unitlength}{\DIFdelgraphicswidth}
				\begin{picture}(1,1)
				\thicklines\linethickness{2pt} 
				{\color[rgb]{1,0,0}\put(0,0){\framebox(1,1){}}}
				{\color[rgb]{1,0,0}\put(0,0){\line( 1,1){1}}}
				{\color[rgb]{1,0,0}\put(0,1){\line(1,-1){1}}}
				\end{picture}
			}\hspace*{3pt}}} 
} 
\LetLtxMacro{\DIFOaddbegin}{\DIFaddbegin} 
\LetLtxMacro{\DIFOaddend}{\DIFaddend} 
\LetLtxMacro{\DIFOdelbegin}{\DIFdelbegin} 
\LetLtxMacro{\DIFOdelend}{\DIFdelend} 
\DeclareRobustCommand{\DIFaddbegin}{\DIFOaddbegin \let\includegraphics\DIFaddincludegraphics} 
\DeclareRobustCommand{\DIFaddend}{\DIFOaddend \let\includegraphics\DIFOincludegraphics} 
\DeclareRobustCommand{\DIFdelbegin}{\DIFOdelbegin \let\includegraphics\DIFdelincludegraphics} 
\DeclareRobustCommand{\DIFdelend}{\DIFOaddend \let\includegraphics\DIFOincludegraphics} 
\LetLtxMacro{\DIFOaddbeginFL}{\DIFaddbeginFL} 
\LetLtxMacro{\DIFOaddendFL}{\DIFaddendFL} 
\LetLtxMacro{\DIFOdelbeginFL}{\DIFdelbeginFL} 
\LetLtxMacro{\DIFOdelendFL}{\DIFdelendFL} 
\DeclareRobustCommand{\DIFaddbeginFL}{\DIFOaddbeginFL \let\includegraphics\DIFaddincludegraphics} 
\DeclareRobustCommand{\DIFaddendFL}{\DIFOaddendFL \let\includegraphics\DIFOincludegraphics} 
\DeclareRobustCommand{\DIFdelbeginFL}{\DIFOdelbeginFL \let\includegraphics\DIFdelincludegraphics} 
\DeclareRobustCommand{\DIFdelendFL}{\DIFOaddendFL \let\includegraphics\DIFOincludegraphics} 

\begin{document}
	
	
	\title{Multimodal Face Synthesis from Visual Attributes} 
	
	\author{Xing Di,~\IEEEmembership{Student Member,~IEEE} and 
		Vishal~M.~Patel,~\IEEEmembership{Senior Member,~IEEE}
		\thanks{Xing Di is with the Whiting School of Engineering, Johns Hopkins University, 3400 North Charles Street, Baltimore, MD 21218-2608, e-mail: xing.di@jhu.edu}
		\thanks{Vishal M. Patel is with the Whiting School of Engineering, Johns Hopkins University, e-mail: vpatel36@jhu.edu}
		\thanks{Manuscript received...}}

	\markboth{Journal of \LaTeX\ Class Files,~Vol.~xx, No.~x, Month~2017}%
	{Shell \MakeLowercase{\textit{et al.}}: Bare Demo of IEEEtran.cls for IEEE Journals}
	%
	
	
	\maketitle

	
	\begin{abstract}
		Synthesis of face images from visual attributes is an important problem in computer vision and biometrics due to its applications in law enforcement and entertainment. Recent advances in deep generative networks have made it possible to synthesize high-quality face images from visual attributes.  However, existing methods are specifically designed for generating unimodal images (i.e visible faces) from attributes. In this paper, we propose a novel generative adversarial network  which simultaneously synthesizes identity preserving multimodal face images (i.e. visible, sketch, thermal, etc.) from visual attributes without requiring paired data in different domains for training the network.  We introduce a novel generator with multimodal stretch-out modules to simultaneously synthesize multimodal face images.  Additionally, multimodal stretch-in modules are introduced in the discriminator which discriminate between real and fake images. Extensive experiments and comparison with several state-of-the-art methods are performed to verify the effectiveness of the proposed attribute-based multimodal synthesis method.
	\end{abstract}

	\section{Introduction}
\begin{figure}[t]
		\centering
		\includegraphics[width=0.9\linewidth]{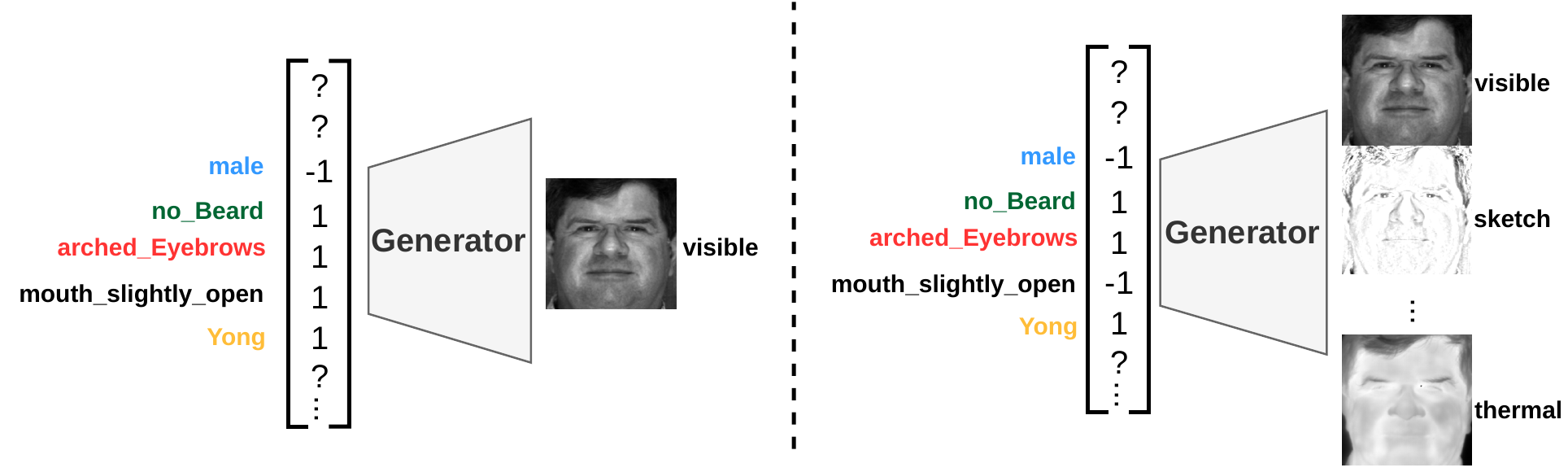}
		\vskip 0pt   (a) Unimodal synthesis \hspace{10mm} (b) Multimodal synthesis
		\vskip -5pt
		\caption{Illustration of unimodal and multimodal face generation from visual attributes. Given a list of facial attributes, we aim to use a single generator to simultaneously synthesize multimodal face images with consistant geometry and texture patterns such that they satisfy these attributes.}
		\label{fig:introduction}
		\vskip -8pt
	\end{figure}

	Generating face images from visual attributes is an important problem in the biometrics and computer vision communities due to its applications in forensics, entertainment, and law enforcement (see Fig.~\ref{fig:introduction}(a)).  For instance, a visual description of a face (i.e. facial attributes) is often used by a forensic artist to render sketch composites of a criminal suspect or a missing person when no facial image of the suspect is available.  Recent advances in deep generative networks have made it possible to generate high-quality face images from facial attributes \cite{yan2016attribute2image,wang2019attribute,zhang2018stackgan++,zhang2017stackgan,zhang2018photographic,di2019facial}. However, existing attribute-to-face synthesis methods mainly focus on generating unimodal face
	images (i.e visible faces) from attributes. In many scenarios, the gallery images contain multiple modalities and the domain gap between the generated unimodal face images and gallery images will degrade the recognition performance. Therefore, a multimodal attribute-to-face synthesis method can assist law enforcement officers to identify a person regardless the domain gap by simultaneously generating face images in visible, sketch and thermal domains.
	
	
	In this paper, we focus on the problem of simultaneously synthesizing multimodal face images from visual attributes (see Fig.~\ref{fig:introduction}(b)).  A naive solution to this problem is to simply use an attribute-to-face synthesis method \cite{yan2016attribute2image, wang2019attribute, zhang2018stackgan++,zhang2017stackgan,zhang2018photographic} to generate a visible image from facial attributes and then use an image-to-image translation method \cite{huang2018multimodal,liu2017unsupervised,zhu2017unpaired,yi2017dualgan,isola2017image} to synthesize images from the visible domain to the other domains such as thermal or composite sketch.  However, it is well-known in the biometrics community that synthesizing facial images from one domain to another (i.e. visible to thermal) itself is an extremely difficult problem and often leads to poor synthesis results \cite{zhang2018synthesis, PhotoSketchLidan,he2019adversarial,yu2019lamp,wu2019disentangled,kong2019cross,zhao2017dual,zhao2018towards,cao2019biphasic,wu2017coupled,fu2019dual,zhao20183d,hu2018pose}.  Hence, a combination of attribute-to-face synthesis with cross-modal synthesis will not be an effective approach for this problem.  
	
	Another approach would be to train $c$ unimodal attribute-to-face synthesis methods separately for $c$ different modalities. However, in this case one often looses geometric and texture consistency among the multimodal face images when the generators are trained separately for each modality. Hence, the synthesized multimodal face images do not contain consistant identities.  Furthermore, training $c$ different networks corresponding to $c$ modalities will require large memory and computation time. 
	
	We propose a new generative adversarial network (GAN), called Att2MFace,  that can directly synthesize  multimodal face images from visual attributes. The generator network consists of  multimodal stretch-out modules which convert the modality invariant features to multimodal images. On the other hand, the  discriminator network contains stretch-in modules which convert the multimodal images to a modality invariant feature representation which can be used to discriminates between real and fake images.  In addition, an  auxiliary estimator is used along with the discriminator to estimate the probability of the target attributes.  By back-propagating the errors of image discriminability and attribute probability, the generator can learn to synthesize a diverse set of realistic multimodal face images from visual attributes.  In addition, we employ  a progressive training strategy \cite{karras2018progressive,karras2019style,karras2020analyzing} to generate photo-realistic high-resolution images.  In particular, we start training our model on a 4$\times$4 resolution and progressively increase it to 256 $\times$ 256 resolution.  Hence, large-scale structure of the image distribution can be first discovered and finer details can then be added by additional layers. 
	
	Extensive experiments and comparison with several state-of-the-art image synthesis methods are performed to verify the effectiveness of the proposed attribute-based multimodal synthesis method. 
	One of the main advantages of the proposed Att2MFace model is that it can generate multimodal images even if there are no paired multimodal images available for training. This is due to the following two reasons: (1) multimodal images are generated from a common feature representation by the stretch-out module, and (2) the texture pattern in lower resolution is inherited to a higher resolution by the progressive-growth training.
	This unsupervised learning setting is convenient for many applications because collecting paired multimodal face images is laborious and expensive. Furthermore, this unsupervised learning shows its benefits in improving the  diversity of generated images \cite{bang2019resembled}. For example, given unpaired visible and sketch images, the generator can explore the unseen texture pattern (complementary information) in the sketch domain while learning to synthesize in the visible domain.	 To summarize, this paper makes the following contributions: 
	
	\begin{itemize}
		
		\item We develop a novel GAN that can simultaneously synthesize multimodal face images from visual attributes.
		
		\item We introduce  multimodal stretch-out and stretch-in modules in the  generator and discriminator networks, respectively.  In addition, a progressive training strategy is employed to generate multimodal photo-realistic high-resolution images with consistant identity.
		
		\item Extensive experiments are conducted to demonstrate the effectiveness of the proposed multimodal image synthesis method.  To the best of our acknowledge, this is the first approach for synthesizing high-quality multimodal face images from  visual attributes.
	\end{itemize}

	
	\section{Related Work}
	\noindent \textbf{Generative Adversarial Networks:} Generative Adversarial Networks  (GANs) \cite{goodfellow2014generative} consist of two parts: a generator $(G)$ and a discriminator $(D)$.   The goal of the generator is to generate new data samples while the discriminator evaluates them for authenticity.  Based on a game theoretic min-max principles, the generator and discriminator networks are optimized jointly by alternating the training of  $D$ and $G$.   The success of  GANs in synthesizing visually appealing images has inspired researchers to explore the use of  GANs in other applications such as multimodal synthesis and face synthesis from attributes.\\
	\noindent \textbf{Multimodal image synthesis:} Various GAN-based multimodal image synthesize methods have been proposed in the literature \cite{choi2018stargan,choi2020stargan,huang2018multimodal,DRIT_plus,DRIT,zhu2017toward,liu2016coupled,mao2018unpaired,yu2019multi,cao2020informative}. For instance, Zhu \etal \cite{zhu2017toward} proposed BicycleGAN for multimodal image synthesis. Huang \etal \cite{huang2018multimodal} introduced the MUNIT model for synthesizing multimodal images by decomposing the input images into content code and style code. Similarly, Lee \etal \cite{DRIT,DRIT_plus} developed the DRIT++ model for  multimodal synthesis by combining the disentangled encoding content feature as well as sampled style features.  Choi \etal \cite{choi2018stargan} proposed StarGAN for multidomain image translation  by adding an extra domain label into the generator. Later on, Choi \etal \cite{choi2020stargan} developed the StarGANv2 model for multidomain image synthesis by mapping noise to different style codes.  Perera \etal \cite{perera2018in2i} proposed an extension of the unsupervised image-to-image translation problem to multiple inputs.  Liu \etal \cite{liu2016coupled} proposed Coupled GAN (CoGAN) for learning a joint distribution of multi-domain images.  Bang \etal \cite{bang2019resembled} developed Resembled-GAN to share the semantic attributes among two domains by adding the feature covariance constraint. Yu \etal \cite{yu2019multi}  introduced an unsupervised  multi-mapping framework which uses the objectives of multi-domain and multi-modal translations for  performing finer image translation.  Mao \etal \cite{mao2018unpaired} introduced RegCGAN to synthesize multimodal images by changing the domain label which is concatenated with the noise vector as input. \\
	\noindent \textbf{Image synthesis from visual attributes:} The problem of synthesizing images conditioned on visual attributes has also gained a lot of traction in recent years due to its application in law enforcement and entertainment. Yan \etal \cite{yan2016attribute2image} proposed a disentangled conditional variational auto-encoding (disCVAE) model for attribute conditioned image generation. Wang \etal \cite{wang2019attribute} proposed a basic conditional GAN (CGAN) method for face synthesis from visual attributes. Recently, Di \etal \cite{di2019facial} developed a cascaded model to synthesize facial images from attributes via sketch. Various face attribute manipulation models have also been proposed in the literature, where the goal is to manipulate face images based on attributes \cite{Wu_2019_ICCV,wu2019relgan,lu2018attribute,dorta2020gan}.
	
	\begin{figure*}[htp!]
		\centering
		\includegraphics[width=0.95\linewidth]{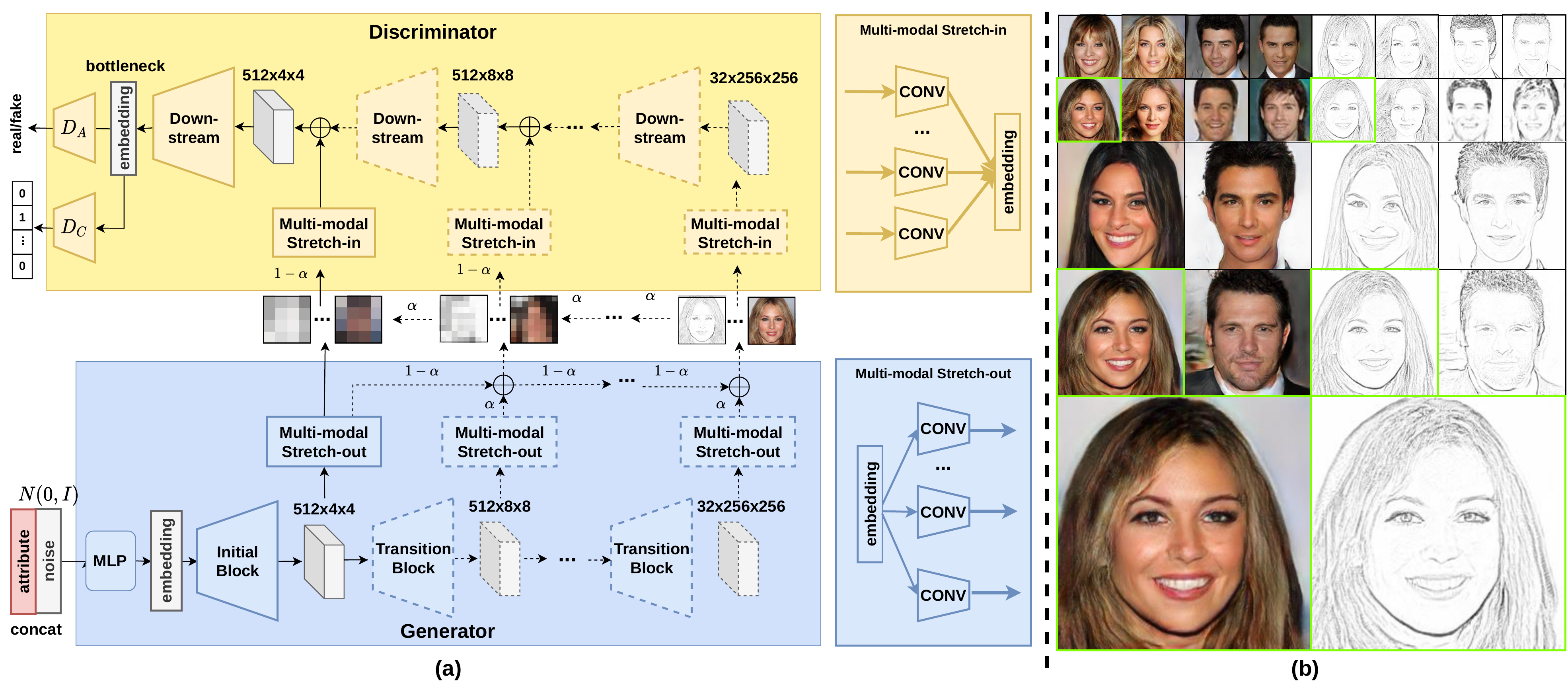} 
		\vskip -8pt
		\caption{An overview of the proposed Att2MFace framework.   (a) Details of the generator and discriminator networks with different modules. The progressive training parts are indicated by dashed lines. (b) Synthesized multimodal images corresponding to the same input at different resolution scales based on progressive training are shown in \textcolor{green}{green boxes}.}
		\label{fig:multimodalgenerator}
	\end{figure*}
	
	\section{Proposed Method} \label{sec:proposed_method}
	In this section, we provide details of the proposed Att2Mface modal, which consists of a multimodal generator and a multimodal discriminator with an auxiliary attribute estimator.   Given a visual attribute vector $\mathbf{y}_{a}$ and a noise vector $\mathbf{z}$ sampled from the normal distribution, the proposed  generator $G$ aims to simultaneously generate multimodal images  $\mathbf{\hat{x}}_{1}, \mathbf{\hat{x}}_{2}, \cdots, \mathbf{\hat{x}}_{c}$ as follows
	\setlength{\belowdisplayskip}{0pt} \setlength{\belowdisplayshortskip}{0pt}
	\setlength{\abovedisplayskip}{0pt} \setlength{\abovedisplayshortskip}{0pt}
	\begin{equation}\label{problem define}
	G(\mathbf{z}, \mathbf{y}_{a} ) = \{\mathbf{\hat{x}}_{1}, \mathbf{\hat{x}}_{2}, \cdots, \mathbf{\hat{x}}_{i}, \cdots, \mathbf{\hat{x}}_{c} \},
	\end{equation}
	where $c$ indicates the total number of face modalities.  
	Fig.~\ref{fig:multimodalgenerator} gives an overview of the proposed network in details.  In what follows, we provide details of the different modules in our framework.
	
	\subsection{MultiModal Generator}
	The proposed generator network consists of the following components: multi-layer perceptron (MLP),  Initial Block,  Transition Blocks and Multi-modal Stretch-out modules.  The MLP module maps $\mathbf{y}_{a}$ and $\mathbf{z}$ to an attribute latent space. MLP consists of one linear fully-connected layer followed by a LeakyReLU function.   The Initial Block learns the feature maps in a particular resolution (i.e. $4\times 4$) from the attribute features.   The Initial Block consists of one reshape function and one convolutional layer followed by a LeakyReLU function.  Then, another Transition Block is implemented to map the feature maps onto a higher resolution (i.e. $8\times 8$) by utilizing an upsampling layer and two convolutional layers. In addition, pixel-wise feature equalization \cite{krizhevsky2017imagenet,karras2018progressive} is employed after all the convolutional layers in the generator.  Pixel-wise feature equalization is defined as follows
	\setlength{\belowdisplayskip}{0pt} \setlength{\belowdisplayshortskip}{0pt}
	\setlength{\abovedisplayskip}{0pt} \setlength{\abovedisplayshortskip}{0pt}
	\begin{equation}\label{eq:feature_equalization}
	\mathcal{F}_{m,n} = \tilde{\mathcal{F}}_{m,n}/\sqrt{\frac{1}{N} \sum_{i=1}^{N} ( \tilde{\mathcal{F}}^{i}_{m,n} )^{2} + \epsilon },
	\end{equation}
	where $\epsilon=10^{-8}$, $N$ is the number of feature maps and $\mathcal{F}_{m,n}$ and $\tilde{\mathcal{F}}_{m,n}$ are the normalized and original feature vectors  in pixel ($m, n$), respectively.  Given a common feature map output from the transition block at a certain resolution scale, a multimodal stretch-out module that contains $c$ convolution operations are implemented to simultaneously generate face images in different modalities. By leveraging the same modality invariant content from the common feature map, this module aims to add modality specific style when converting the feature maps to a output images.

	\subsection{MultiModal Discriminator} \label{sec:Multi-modal Discriminators}
	As shown in Fig.~\ref{fig:multimodalgenerator}, the multimodal discriminator, $D$ contains two output streams: estimation $D_{C}$ and  authentication $D_{A}$. The authentication stream, $D_{A}$, measures the probability of whether a given multimodal sample belongs to the data distribution. On the other hand, the estimation stream, $D_{C}$, aims to learn the mapping from the input image to the corresponding target label probability distribution, which is achieved by adding an auxiliary estimator.  This estimator allows a single discriminator to learn the discriminability based on the corresponding multimodal image labels \cite{odena2017conditional}.

	The discriminator contains a series of Multimodal Stretch-in and Down-stream modules.  For each resolution scale, a multimodal stretch-in module is employed which contains $c$ convolution operators followed by  LeakyReLU. This module distills the modality specific style and converts the three-channel input to modality invariant multi-channel representation. Given the feature maps from the multi-modal stretch-in module, the Down-stream module learns a representation at different resolutions.  The Down-stream module consists of one average pooling layer and two convolution layers followed by feature equalization. Finally, after a series of Down-stream blocks, the bottleneck feature maps are of size 4$\times$4 and are passed on to both the estimation stream, $D_{C}$ and the authentication steam, $D_{A}$. The authentication and estimation streams contain two fully-connected layers to learn the probability of discrimination and the corresponding target label, respectively.

	To make sure that the estimator learns both the modality and attribute probabilities, we define a new target label $\mathbf{y}$ as follows:
	\setlength{\belowdisplayskip}{0pt} \setlength{\belowdisplayshortskip}{0pt}
	\setlength{\abovedisplayskip}{0pt} \setlength{\abovedisplayshortskip}{0pt}
	\begin{equation}\label{eq:new target label}
	\mathbf{y}_{i} = [ \mathbf{y}_{a} , \mathbf{y}_{m} ],\;\; \text{where } \;\;\mathbf{y}_{m} = [0,1,\cdots,0],
	\end{equation}   
	where, $\mathbf{y}_{a}$ is the visual attribute vector, $\mathbf{y}_{m} = [0,1,\cdots,0]$ is a $c$-dimensional one-hot vector indicating what modality the image belongs to, and $c$ is the total number of modalities.

	\subsection{Progressive Training}
	
	Recent image generation approaches have found  progressive training beneficial \cite{karras2018progressive,karras2019style,karras2020analyzing}. In progressive training, the idea is to start with low-resolution images and then progressively increase the resolution by adding layers to the network.  This way, large-scale structure of the image distribution is first discovered and then finer details are added by additional layers.  Hence, we start training our model from resolution $4\times4$ and progressively grow it to $8\times8, \cdots, 256\times256$.  To achieve this, skip connection between the output of newly added Transition/Down-stream Block and the existing output/input of the  generator/discriminator are balanced by a trainable weight parameter $\alpha$.   Starting from a light weight $\alpha$ ($\alpha=0$) helps to smooth out  the influence of adding a new Transition/Down-stream block to the network. Additionally, the equalization operations as defined in Eq.~\ref{eq:feature_equalization} are used in both generator and discriminator networks to prevent the escalation of feature magnitudes.

	\subsection{Objective Function} \label{sec:multimodal objective}
	The objective functions used to train the multimodal generator and the discriminator networks consist of the adversarial loss and the classification loss for authentication and estimation streams, respectively.
	
	\noindent \textbf{Adversarial Loss:} In order to make the generated multimodal images indistinguishable from real multimodal images, we adopt the WGAN-GP loss \cite{gulrajani2017improved,arjovsky2017wasserstein} which is defined as follows:
	\begin{equation}\label{eq:adversarial loss}
	\begin{split}
	\mathcal{L}_{adv} = \sum_{i=1}^{c} (\mathbb{E}[D_{A}(\mathbf{x}_{i})] - \mathbb{E}[ (D_{A}(\mathbf{\hat{x}}_{i}))] - \\ 
	\lambda_{gp} \mathbb{E}[(\|\triangledown_{\mathbf{x}^{\star}} D_{A}(\mathbf{x}_{i}^{\star}) \|_{2} -1)^{2}]),
	\end{split}
	\end{equation}
	where $\mathbf{x}_{i}$ and $\mathbf{\hat{x}}_{i}$ correspond to the real image and the synthesized image corresponding to the $i$-th modality, respectively.  Here, $\mathbf{x}_{i}^{\star}$ is sampled uniformly along a straight line between a pair of real $\mathbf{x}_{i}$ and the generated $\mathbf{\hat{x}}_{i}$ \cite{choi2018stargan}.  $D_A$ refers to the output probability score from the authentication stream.  The discriminator attempts to maximize this objective, while the generator attempts to minimize it.  We set $\lambda_{gp}=10$ in our experiments.\\  
	\noindent \textbf{Classification Loss:} The classifier $D_{C}$ aims to learn the mapping from the input images to the target label probability distribution. On the other hand, the generator $G$ aims to synthesize images that match the corresponding target label.  In order to achieve this, the classification loss is imposed when optimizing both $D_{C}$ and $G$. The objective consists of two separate terms: (i) the classifier $D_{C}$ is optimized by the real images $\mathbf{x}_{i}$ and the target labels $\mathbf{y}_{i}$, and (ii) the generator $G$ is optimized to synthesize $\mathbf{\hat{x}}_{i}$ with corresponding target label $\mathbf{y}_{i}$.  In particular, the first term is defined as follows:
	\begin{equation}\label{eq:real classification loss}
	\mathcal{L}_{cls}^{real} = \sum_{i=1}^{c} \mathbb{E}_{\mathbf{x}_{i}, \mathbf{y}_{i}}[-\log P( D_{C}(\mathbf{x}_{i})= \mathbf{y}_{i} | \mathbf{x}_{i}) ],
	\end{equation}
	where $D_{C}(\mathbf{x}_{i})$ denotes the class probability vector over the classification module $D_{C}$ given real input $\mathbf{x}_{i}$. By minimizing this objective with the target label $\mathbf{y}_{i}$, the discriminator learns to classify a real image to its correct target label. For second term, the loss function is defined as follows
	\begin{equation}\label{eq:fake classification loss}
	\mathcal{L}_{cls}^{fake} = \sum_{i=1}^{c} \mathbb{E}_{\mathbf{\hat{x}}_{i}, \mathbf{y}_{i}}[-\log P (D_{C}(\mathbf{\hat{x}}_{i})=\mathbf{y}_{i} | \mathbf{\hat{x}}_{i})],
	\end{equation}
	where $\mathbf{\hat{x}}_{i} \sim G(\mathbf{z}, \mathbf{y}_{a})_{i}$ is the $i$-th modality fake image. The generator $G$ attempts to minimize this classification objective so that the synthesized image has the corresponding target label $\mathbf{y}_{i}$.\\
	\noindent \textbf{Overall Objective:}
	The overall objective function to optimize the discriminator $D$ and the generator $G$ is defined as follows
	\begin{equation}\label{full objective}
	\begin{split}
	\mathcal{L}_{D} = -\mathcal{L}_{adv} + \lambda_{cls}\mathcal{L}_{cls}^{real} \\
	\mathcal{L}_{G} = \mathcal{L}_{adv} + \lambda_{cls}\mathcal{L}_{cls}^{fake},
	\end{split}
	\end{equation}
	where $\lambda_{cls}$ is the parameter that controls the contribution of the classification losses. In our experiments, we have found that a larger value of  $\lambda_{cls}$ leads to a slower training convergence and a smaller value of $\lambda_{cls}$ does not make the synthesized images preserve the corresponding attributes well.  We set $\lambda_{cls}=1$ in our experiments as it gives us  better performance.
	
	During training, the generator and discriminator networks are optimized iteratively. When updating the discriminator, $D$ is optimized by maximizing the difference from synthetic data to real data distribution (authentication $D_{A}$).  On the other hand, it is also optimized by estimating the corresponding ground-truth target label (estimation $D_{C}$). When updating the generator, $G$ is optimized by minimizing the difference from the synthetic data to real data distribution. It is also optimized by synthesizing data samples that match the corresponding target label. In this way, the synthesized images are not only photo-realistic but also satisfy the correct target label.	
	
	\begin{table}[t]
		\caption{The generator and discriminator network architectures that we use to generate 256x256 multimodal images.}\label{tab:network architecture}
		\begin{minipage}{1.0\columnwidth}
			\centering
			\resizebox{0.95\columnwidth}{!}{%
				\begin{tabular}{|c|c|c|c|}
					\hline 
					Generator & Act. & Output shape & Block  \\ 
					\hline 
					input code &  & $ (512 + d_{a})$  & - \\ 
					Fully-connected &  LReLU, Reshape & $ 512 \times 4 \times 4 $ & MLP \\
					Conv $3\times 3$ & LReLU & $512 \times 4 \times 4$ & Initial Block \\
					\hline
					$c \times$ Conv $1\times 1$ & & $3 \times 4 \times 4$ & Multimodal Stretch-out \\
					\hline
					Upsample & & $512 \times 8 \times 8$ & \multirow{3}{*}{Transition Block} \\
					Conv $3\times 3$ & LReLU & $512 \times 8 \times 8$ &  \\
					Conv $3\times 3$ & LReLU & $512 \times 8 \times 8$ &  \\
					\hline 
					$c \times$ Conv $1\times 1$ & linear & $3 \times 8 \times 8$ & Multimodal Stretch-out \\
					\hline
					Upsample & & $512 \times 16 \times 16$ & \multirow{3}{*}{Transition Block} \\
					Conv $3\times 3$ & LReLU & $256 \times 16 \times 16$ &  \\
					Conv $3\times 3$ & LReLU & $256 \times 16 \times 16$ &  \\
					\hline 
					$c \times$ Conv $1\times 1$ & linear & $3 \times 16 \times 16$ & Multimodal Stretch-out \\
					\hline
					Upsample & & $256 \times 32 \times 32$ & \multirow{3}{*}{Transition Block} \\
					Conv $3\times 3$ & LReLU & $128 \times 32 \times 32$ &  \\
					Conv $3\times 3$ & LReLU & $128 \times 32 \times 32$ &  \\
					\hline
					$c \times$ Conv $1\times 1$ & linear & $3 \times 32 \times 32$ & Multimodal Stretch-out \\
					\hline
					Upsample & & $128 \times 64 \times 64$ & \multirow{3}{*}{Transition Block} \\
					Conv $3\times 3$ & LReLU & $64 \times 64 \times 64$ &  \\
					Conv $3\times 3$ & LReLU & $64 \times 64 \times 64$ &  \\
					\hline 
					$c \times$ Conv $1\times 1$ & linear & $3 \times 64 \times 64$ & Multimodal Stretch-out \\
					\hline 
					Upsample & & $64 \times 128 \times 128$ & \multirow{3}{*}{Transition Block} \\
					Conv $3\times 3$ & LReLU & $32 \times 128 \times 128$ &  \\
					Conv $3\times 3$ & LReLU & $32 \times 128 \times 128$ &  \\
					\hline 
					$c \times$ Conv $1\times 1$ & linear & $3 \times 128 \times 128$ & Multimodal Stretch-out \\
					\hline 
					Upsample & & $32 \times 256 \times 256$ & \multirow{3}{*}{Transition Block} \\
					Conv $3\times 3$ & LReLU & $16 \times 256 \times 256$ &  \\
					Conv $3\times 3$ & LReLU & $16 \times 256 \times 256$ &  \\
					\hline 
					$c \times$ Conv $1\times 1$ & linear & $3 \times 256 \times 256$ & Multimodal Stretch-out \\
					\hline 
				\end{tabular} 
			}
		\end{minipage}%
		
		\begin{minipage}{1.0\columnwidth}
			\centering
			\resizebox{0.95\columnwidth}{!}{%
				\begin{tabular}{|c|c|c|c|}
					\hline 
					Discriminator & Act. & Output shape & Block  \\ 
					\hline 
					input images &  & $  3 \times 256 \times 256$  & - \\
					\hline
					$c \times$ Conv $1\times 1$ & linear & $ 16 \times 256 \times 256 $ & Multimodal Stretch-in \\
					\hline
					Conv $3\times 3$ & LReLU & $16 \times 256 \times 256$ & \multirow{3}{*}{Down-stream} \\
					Conv $3\times 3$ & LReLU & $32 \times 256 \times 256$ &  \\
					Downsample & & $32 \times 128 \times 128$ & \\
					\hline
					$c \times$ Conv $1\times 1$ & linear & $32 \times 128 \times 128$ & Multimodal Stretch-in \\
					\hline
					Conv $3\times 3$ & LReLU & $32 \times 128 \times 128$ & \multirow{3}{*}{Down-stream} \\
					Conv $3\times 3$ & LReLU & $64 \times 128 \times 128$ &  \\
					Downsample & & $64 \times 64 \times 64$ & \\
					\hline			
					$c \times$ Conv $1\times 1$ & linear & $64 \times 64 \times 64$ & Multimodal Stretch-in \\
					\hline
					Conv $3\times 3$ & LReLU & $64 \times 64 \times 64$ & \multirow{3}{*}{Down-stream} \\
					Conv $3\times 3$ & LReLU & $128 \times 64 \times 64$ &  \\
					Downsample & & $128 \times 32 \times 32$ & \\
					\hline					
					$c \times$ Conv $1\times 1$ & linear & $128 \times 32 \times 32$ & Multimodal Stretch-in \\
					\hline
					Conv $3\times 3$ & LReLU & $128 \times 32 \times 32$ & \multirow{3}{*}{Down-stream} \\
					Conv $3\times 3$ & LReLU & $256 \times 32 \times 32$ &  \\
					Downsample & & $256 \times 16 \times 16$ & \\
					\hline	
					$c \times$ Conv $1\times 1$ & linear & $256 \times 16 \times 16$ & Multimodal Stretch-in \\
					\hline
					Conv $3\times 3$ & LReLU & $256 \times 16 \times 16$ & \multirow{3}{*}{Down-stream} \\
					Conv $3\times 3$ & LReLU & $512 \times 16 \times 16$ &  \\
					Downsample & & $512 \times 8 \times 8$ & \\
					\hline
					$c \times$ Conv $1\times 1$ & linear & $512 \times 8 \times 8$ & Multimodal Stretch-in \\
					\hline
					Conv $3\times 3$ & LReLU & $512 \times 8 \times 8$ & \multirow{3}{*}{Down-stream} \\
					Conv $3\times 3$ & LReLU & $512 \times 8 \times 8$ &  \\
					Downsample & & $512 \times 4 \times 4$ & \\
					\hline	
					$c \times$ Conv $1\times 1$ &  & $512 \times 4 \times 4$ & Multimodal Stretch-in \\
					\hline
					Conv $3\times 3$ & LReLU & $512 \times 4 \times 4$ & \multirow{2}{*}{Down-stream} \\
					& Reshape & $8192$ &  \\
					\hline
					Fully-connected / Fully-connected &  & $16$ / ($d_{a} + d_{c}$) & $D_{A}$ / $D_{C}$ \\
					\hline		
				\end{tabular}	
			}
		\end{minipage}
	\end{table}

	\subsection{Network Architecture}
	
	The architectures corresponding to the multimodal generator and multimodal discriminator are shown in Table~\ref{tab:network architecture}. The values the noise code $\mathbf{z} \in \mathcal{R}^{512}$ and the attribute vector $\mathbf{y}_{a} \in \mathcal{R}^{d_{a}}$ take are in the range $[-1, 1]$. Both networks consist of various Transition and Down-stream Blocks. Both the Multimodal Stretch-out and Stretch-in modules consist of $c$ numbers of $1\times 1$ convolutional layers corresponding to  $c$ different modalities.  Except for the stretch-out/stretch-in module on the $4\times 4$ resolution, the multimodal output images are linearly interpolated with the outputs from the Conv $1\times 1$ layer with the upsampled/downsampled ones from the former resolution scale by the trainable weight $\alpha$. 
	
	We use the Leaky ReLU operation with leakiness 0.2 in all the layers of both generator and discriminator. Besides, we use  pixel-wise equalization instead of batch normalization or instance normalization after each Conv $3 \times 3$ layer in both networks.  Upsample and Downsample operations utilize $2 \times 2$ element replication and average pooling, respectively. At the end of the discriminator, two fully-connected layers are utilized for authentication $D_{A}$ and classification $D_C$ individually. The output neurons of these two fully-connected layers are trained with the WGAN-GP and cross-entropy loss separately.

	For the other baseline networks, we use the noise code and the attribute code with the same dimension. Additionally,  we follow the training configuration described in those respective papers as closely as
	possible. For CoGAN, we extend the number of output branches to $c$ modalities by replication. The classifier is adopted with the cross-entropy loss. Except for the fist layer of the discriminator and the last layer of the generator, the rest of layers are all with the tied-weights. For RegCGAN, we extend the domains from 2 in the original paper to $c$. We obtain the multimodal face images during testing by changing the domain label while keeping the noise and attribute code the same.

	\section{Experiments} \label{sec:experimental_results}

	In this section, the experimental evaluations of the proposed method are discussed in detail.  We compare Att2MFace against several multimodal synthesis baselines both qualitatively and quantitatively. In addition, we present attribute and noise manipulation results.

	\begin{figure}
		\centering
		\includegraphics[width=0.95\linewidth]{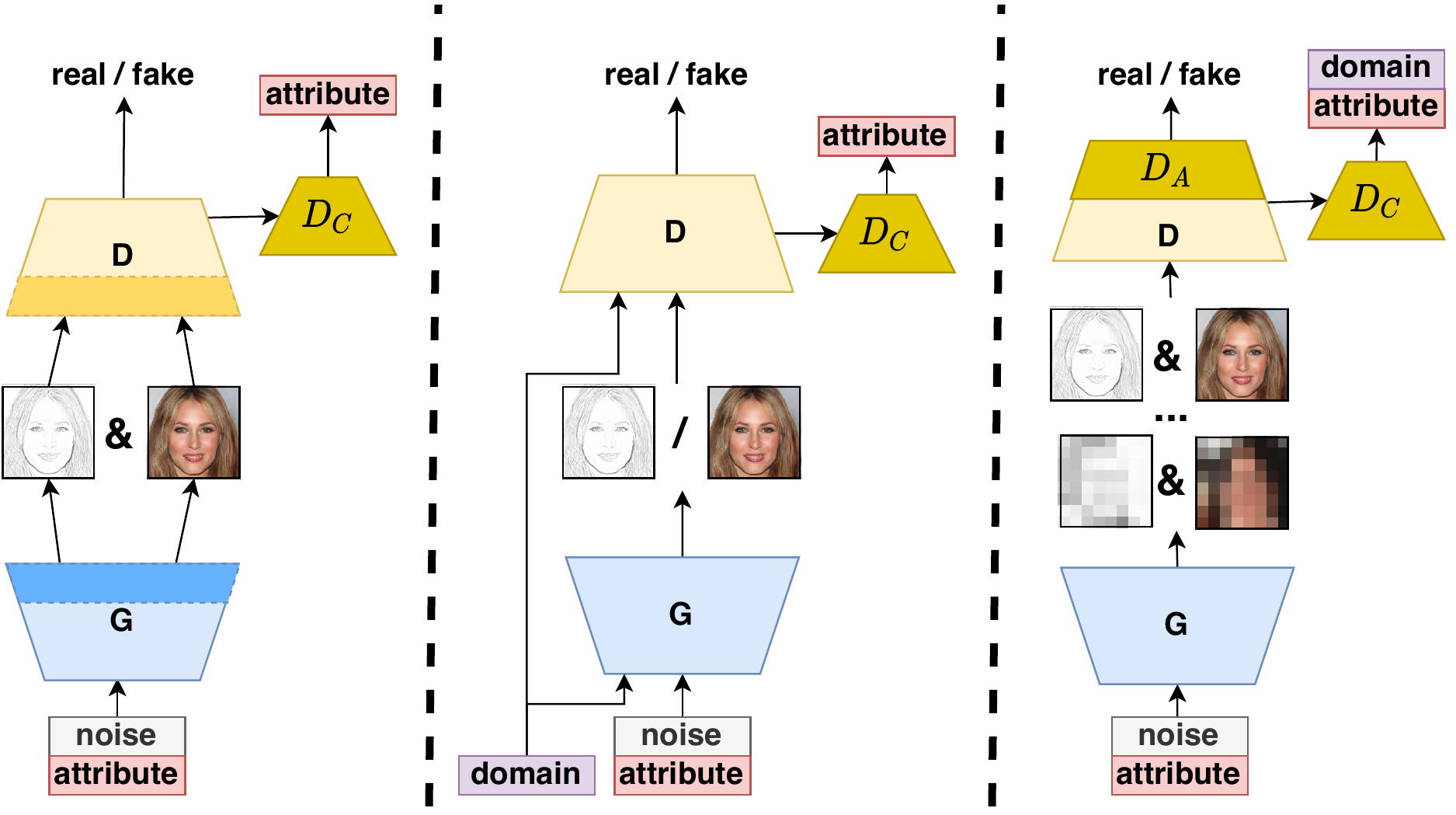} \\
		\hskip -4pt (a) CoGAN\cite{liu2016coupled} \hskip 12pt (b) RegCGAN\cite{mao2018unpaired} \hskip 12pt (c) Att2MFace \\
		\caption{Comparison among baseline models.}
		\label{fig:comparisonbaselines}
		\vskip -12pt
	\end{figure}
	
	\subsection{Baseline Models}
	We consider the following two recently proposed multimodal image synthesis methods as baselines: CoGAN \cite{liu2016coupled} and RegCGAN \cite{mao2018unpaired}.  Both of these methods perform noise-to-image synthesis over multiple modalities.  We concatenate visual attributes with noise to generate attribute conditioned multimodal images.  In another baseline, we use an output from an attribute-to-face synthesis network as an input to multimodal image-to-image translation network such as StarGANv2 \cite{choi2020stargan}.  Fig.~\ref{fig:comparisonbaselines} presents a comparison among these baseline methods. In what follows, we give more details regarding these baselines.\\
	\noindent {\bf{CoGAN \cite{liu2016coupled}}} utilizes coupled generators/ discriminators that share weights in shallow/deeper layers to synthesize multimodal images. The generators and discriminators are jointly optimized. We adopt the architecture of CoGAN with a classifier to predict the attribute.  Except for the last layer of the generator and the first layer of the discriminator, the rest of the layers from each modality network are tied.  \\
	\noindent {\bf{RegCGAN \cite{mao2018unpaired}}}  generates multimodal images by changing the modality label $\mathbf{y}_{m}$.  Training this model is regularized by forcing the first layer generator features to be similar for paired inputs and forcing the last layer features of the discriminator to be similar for paired inputs. Similar to CoGAN, an auxiliary classifier is attached to the discriminator for predicting the attributes. \\
	\noindent {\bf{StarGANv2$^{\star}$ \cite{choi2020stargan}}} is the state-of-the-art multimodal image-to-image translation model. We re-trained StarGANv2 based on different modalities available in each dataset. For a fair comparison, visible images from our Att2MFace network are used as input to StarGANv2. This network synthesizes the remaining multimodal images from the given visible image.

	\begin{table*}[htp!]
		\centering
		\caption{Quantitative results in terms of the FID $\downarrow$  (LPIPS $\uparrow$) scores corresponding to different methods. Mean value is calculated by averaging the available FID scores from each modality. }
		\label{tab: quantitative_fid_result}
		\resizebox{2.0\columnwidth}{!}{%
			\begin{tabular}{|c|c|c|c|c|c|c|c|c|c|c|}
				\hline
				& \multicolumn{4}{c|}{ARL Multimodal Face Database}          & \multicolumn{3}{c|}{CelebA Database} & 		 \multicolumn{3}{c|}{CASIA NIR-VIS 2.0} \\ \hline\hline
				Methods & Visible  & Polarimetric & S0 & Mean & Visible & Sketch & Mean & Vis & NIR & Mean \\ \hline
				
				CoGAN \cite{liu2016coupled} & 397.22 (0.3942) & 275.60 (\textbf{0.4647}) & 311.55 (\textbf{0.4870}) & 328.12 (0.4486) & 110.92 (0.5053) & 113.29 (0.3222) & 112.10 (0.4137) & 306.03 (0.3103) & 97.27 (\textbf{0.3494}) & 201.65 (0.3298) \\ \hline
				
				RegCGAN \cite{mao2018unpaired} & 382.75 (0.4792) & 264.86 (0.4484) & 299.53 (0.4079) & 315.71 (0.4451) & 108.55 (0.5183) & 118.52 (0.3338) & 113.53 (0.4260) & 142.32 (0.3201) & 117.66 (0.3467) & 129.99 (0.3334) \\ \hline
				
				StarGANv2$^{\star}$ \cite{choi2020stargan} & 65.26 (0.4848) & 127.17 (0.1078) & 157.54 (0.1684) & 116.65 (0.2536) & 13.30 (0.5494) & 115.51 (0.2039) & 64.40 (0.3766) & 35.03 (0.3815) & 54.45 (0.1707) & 44.74 (0.2761) \\ \hline
				
				Att2MFace (ours) & \textbf{65.26 (0.4848)} & \textbf{43.04} (0.4542) & \textbf{69.36} (0.4351) & \textbf{59.22 (0.4580)} & \textbf{13.30 (0.5494)} & \textbf{17.75 (0.4568)} & \textbf{15.52 (0.5031)} & \textbf{35.03 (0.3815)} & \textbf{52.64}  (0.3343) & \textbf{43.83 (0.3579)} \\ \hline
				
			\end{tabular}
		}
	\end{table*}
	
	\subsection{Datasets}
\begin{figure}[t]
		\centering
		\includegraphics[width=1.0\linewidth]{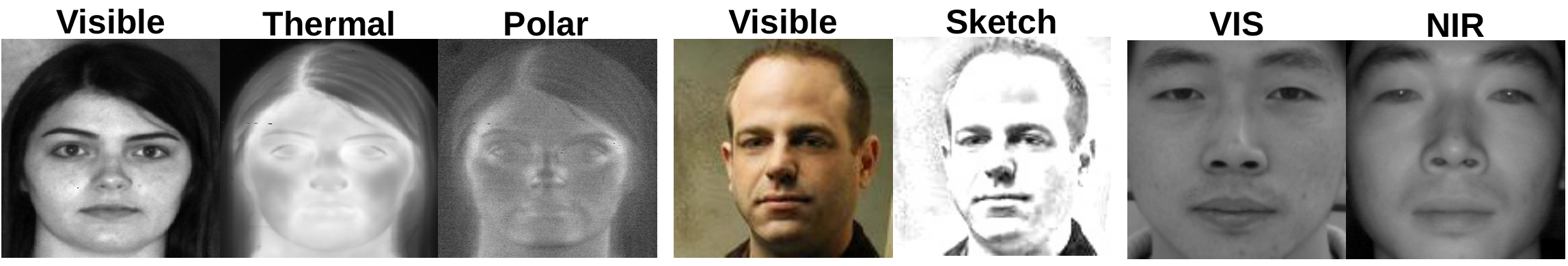}\\
		\raggedright \hspace{40pt} (a) \hspace{68pt}  (b) \hspace{56pt}  (c)
		\vskip -6pt
		\caption{Sample images and the corresponding modalities from  (a) ARL Multi-modal Face Dataset \cite{zhang2018synthesis,hu2016polarimetric}, (b) CelebA-HQ \cite{karras2018progressive}, and (c) CASIA NIR-VIS 2.0 \cite{li2013casia}.}
		\label{fig:datasets}
	\end{figure}

	The following three multimodal face datasets are used to conduct experiments: ARL Multi-modal Face Dataset \cite{hu2016polarimetric,zhang2018synthesis}, CelebA-HQ dataset \cite{karras2018progressive} and CASIA NIR-VIS 2.0 Face Database \cite{li2013casia}. 	Sample images from different modalities from these datasets are shown in Fig.~\ref{fig:datasets}.  The list of facial attributes used from these datasets to synthesize multimodal face images is tabulated in Table~\ref{tab:selected attributes}. During training, multimodal images are randomly sampled according to the attributes.  Hence, multimodal data are not required in pairs.
	
	\begin{table}[t]
		\centering
		\small
		\caption{List of selected visual-attributes.}\label{tab:selected attributes} 
		\resizebox{1.0\columnwidth}{!}{%
			\begin{tabular}{|c|c|}
				\hline 
				ARL Multimodal Face Dataset &  \makecell{ Male, Mouth\_Open, \\ Mustache, No\_Bear and Young.} \\		
				\hline  CelebA-HQ  & \makecell{ Male, Young, Smiling, \\ Mouth\_Open, No\_Beard.} \\
				\hline  CASIA NIR-VIS 2.0 &  Male, Eye\_Glasses, Smiling \\		
				\hline
			\end{tabular}
		}
	\end{table}

	\noindent\textbf{ARL Multimodal Face Dataset:}
	The ARL multimodal dataset \cite{hu2016polarimetric,zhang2018synthesis} consists of 5,419 polarimetric thermal and visible pairs of images from 121 subjects in various expressions, pose, etc. conditions.  We resize the images to 256$\times$256 resolution for training our model.  
	
	
	
	\noindent \textbf{CelebA-HQ:}  The CelebA-HQ dataset \cite{karras2018progressive} consists of 30,000 high-resolution face images (i.e. 1024$\times$ 1024) with  corresponding 40 facial attributes \cite{liu2015faceattributes}.  We extract sketch images \footnotemark[2] from the visible images and view them as the second modality images.  We resize the images to 256$\times$256 resolution for training our model.

	
	\footnotetext[2]{http://www.askaswiss.com/2016/01/how-to-create-pencil-sketch-opencv-python.html}

	\noindent \textbf{CASIA NIR-VIS 2.0:}  The CASIA NIR-VIS 2.0 Face Dataset \cite{li2013casia} contains 12,487 near-infrared (NIR) images and 5,093 visible images corresponding to 725 subjects.  The images in this dataset have been tightly cropped with 128$\times$128 resolution.
	

	\subsection{Implementation}
	The Adam optimizer \cite{kingma2014adam} with a batch size of 16 is used to train the network. The learning rate starts from 0.001 for the generator and the discriminator. The number of iteration for seven scales $4\times 4, 8\times 8, \cdots, 256\times 256$ are set equal to $4.8 \times 10^{4}, 9.6 \times 10^{4}, \cdots 9.6 \times 10^{4}, 2.0 \times 10^{5}$, respectively. It takes around 8 days to train the entire network on two NVIDIA TITAN Xp GPUs. The hyperparameters are selected based on the lowest FID score from 5000 random samples during training.
	
	In CoGAN \cite{liu2016coupled}, the multimodal face images are synthesized by concatenating the noise vector with visual attributes. In RegCGAN \cite{mao2018unpaired}, different modality images are obtained by feeding noise and attribute concatenated vector with a specific modality label $c$ different times.  Regarding StarGANv2$^{\star}$ \cite{choi2020stargan}, the model is first re-trained based on particular modalities and then the synthesized visible image from Att2MFace is used as input to generate multimodal images corresponding to other modalities.

	\subsection{Results}
	We compare our method with the baseline models qualitatively and quantitatively.	The performance of different methods is quantitatively evaluated using the Fr$\acute{\text{e}}$chet inception distance (FID) \cite{heusel2017gans} and Learned Perceptual Image Patch Similarity (LPIPS) distance \cite{zhang2018unreasonable}. A lower FID value implies that the generated data are closer to the real data. When computing the FID score, we choose the same number of synthetic samples as the number of original images in the related dataset. Higher LPIPS values imply that the synthesized images are more diverse.  Following \cite{zhu2017toward,lee2018diverse,huang2018multimodal}, we randomly choose 2k synthetic images and measure the pairwise LPIPS for each modality.

	\begin{figure*}
		\begin{minipage}[c]{\textwidth}
			\centering
			\includegraphics[width=1.0\linewidth]{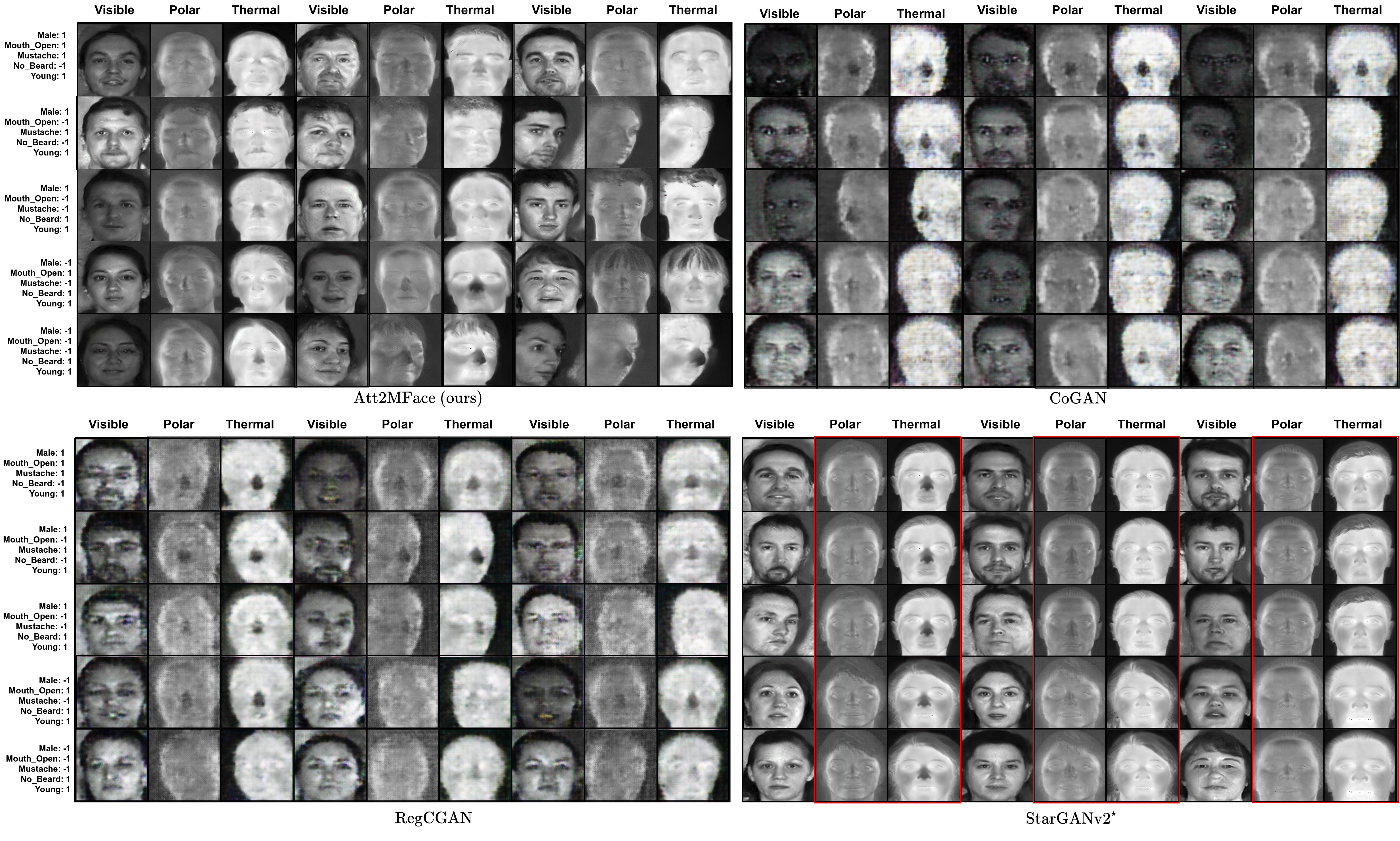}
			\vskip -12pt
			\caption{Sample 256 $\times$ 256 resolution multimodal images generated by different methods using the ARL Multimodal Face Database (Zoom-in for better visualization). The \textcolor{red}{red boxes} highlight the loss of  geometry consistency by StarGANv2.}
			\label{fig:arl256x256samples}
		\end{minipage}
		\begin{minipage}[c]{\textwidth}
			\centering
			\includegraphics[width=1.0\linewidth]{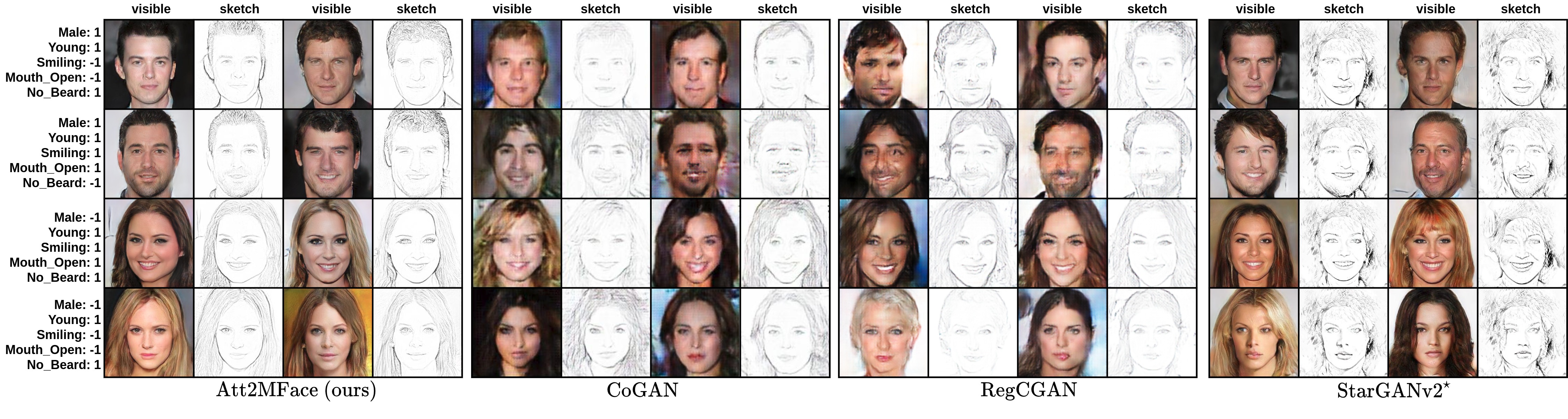}
			\vskip -12pt
			\caption{Sample 256 $\times$ 256 resolution multimodal images generated by different methods using the CelebA-HQ dataset. (Zoom-in for better visualization)}
			\label{fig:celebahq256x256samples}
		\end{minipage}
		\begin{minipage}[c]{\textwidth}
			\centering
			\includegraphics[width=1.0\linewidth]{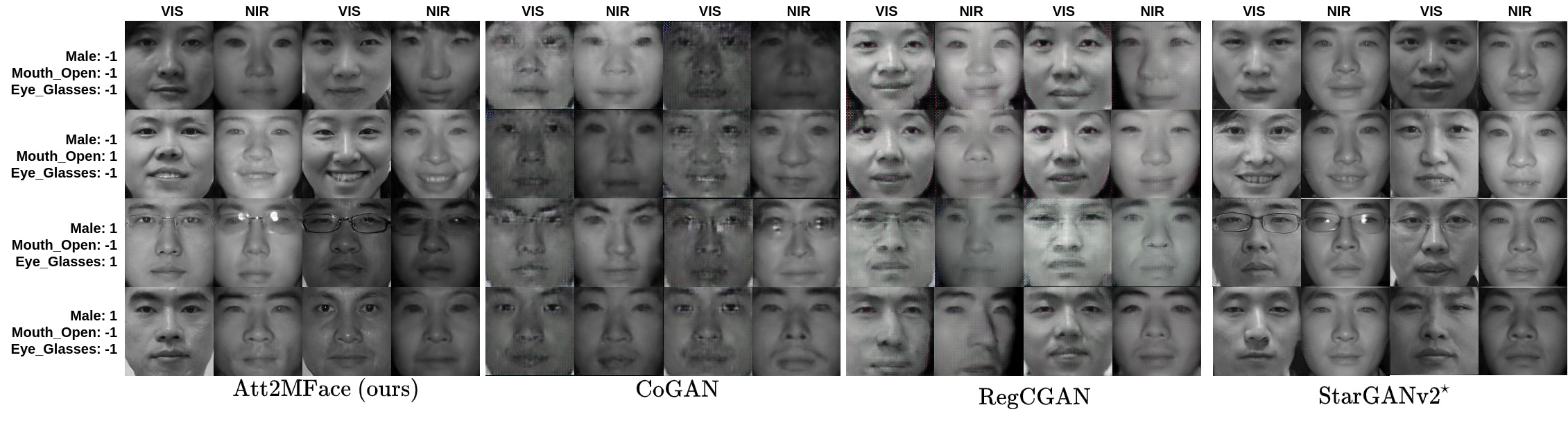}
			\vskip -12pt
			\caption{Sample 128 $\times$ 128 resolution multimodal images generated by different methods using the CASIA NIR-VIS 2.0 dataset. (Zoom-in for better visualization)}
			\label{fig:casia128x128samples}
		\end{minipage}
	\end{figure*}

	\noindent {\bf{Image Quality:}} The quantitative results corresponding to different methods are shown in Table~\ref{tab: quantitative_fid_result}.  Figs.~\ref{fig:arl256x256samples}, \ref{fig:celebahq256x256samples}, \& \ref{fig:casia128x128samples} show the qualitative performance of different methods on the ARL Multi-modal Face Dataset,  CelebA-HQ and CASIA NIR-VIS 2.0 datasets, respectively.  As shown in these figures, Att2MFace is able to synthesize multimodal images directly from visual attributes better than the other methods. In particular, the generated images preserve the attributes that were used to synthesize images (listed on the first column). Regarding the other baseline methods, CoGAN \cite{liu2016coupled} and RegCGAN \cite{mao2018unpaired} are also able to produce reasonable images but the image quality from our model is more photo-realistic than those two methods.  Furthermore, the LPIPS distances corresponding to Att2MFace are higher than the other baseline methods which indicates that the generated images from our method are more diverse.  The performance of StarGANv2$^{\star}$ depends on the quality of the input image which is synthesized by Att2MFace.  It also depends on how well StarGANv2$^{\star}$ is able to synthesize multimodal images from the visible image.  As can be seen from these Figs.~\ref{fig:arl256x256samples}, \ref{fig:celebahq256x256samples}, \& \ref{fig:casia128x128samples}  and Table~\ref{tab: quantitative_fid_result}, in general, StarGANv2$^{\star}$ can produce multimodal images but the image quality is poor and the diversity is limited because of the domain discrepancy among different modalities.  This experiment clearly shows the significance of our multimodal image generation method.

\begin{figure*}[htp!]
		\centering
		\includegraphics[width=0.9\linewidth]{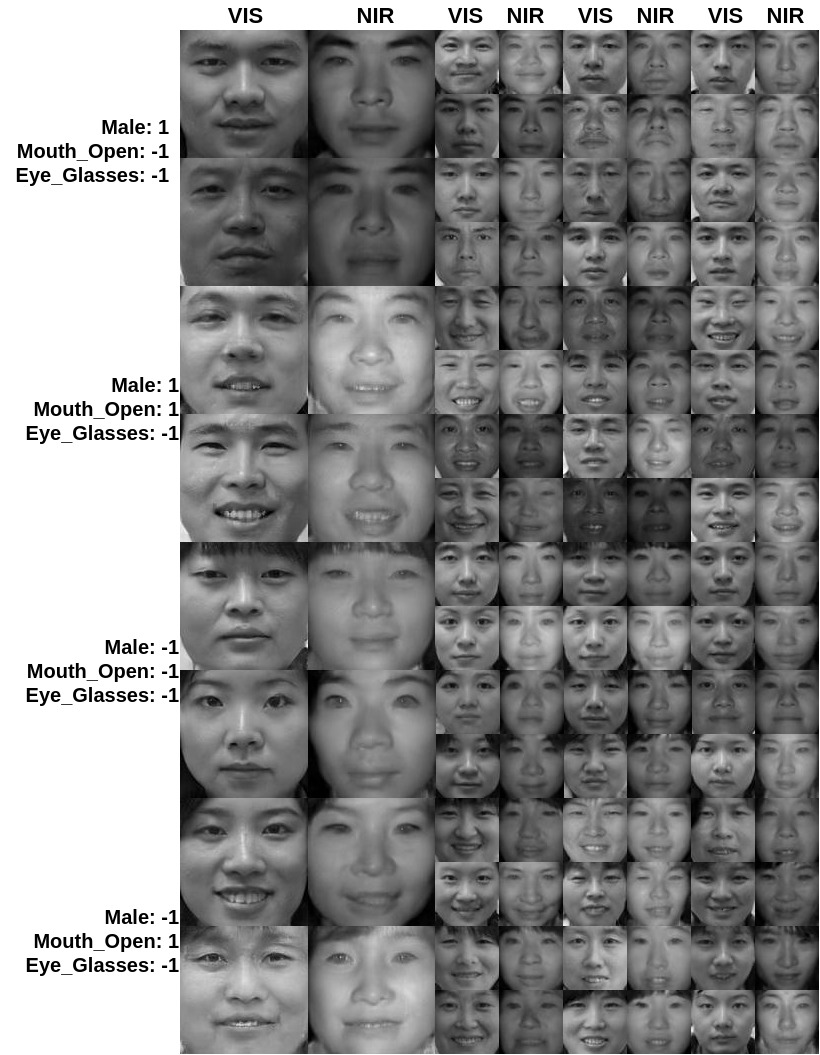}
		\caption{Additional 128x128 and 64x64 multimodal images generated using the CASIA-NIR-VIS 2.0 dataset.}
		\label{fig:additional_Casia}
	\end{figure*}

\begin{figure*}[htp!]
		\centering
		\includegraphics[width=0.9\linewidth]{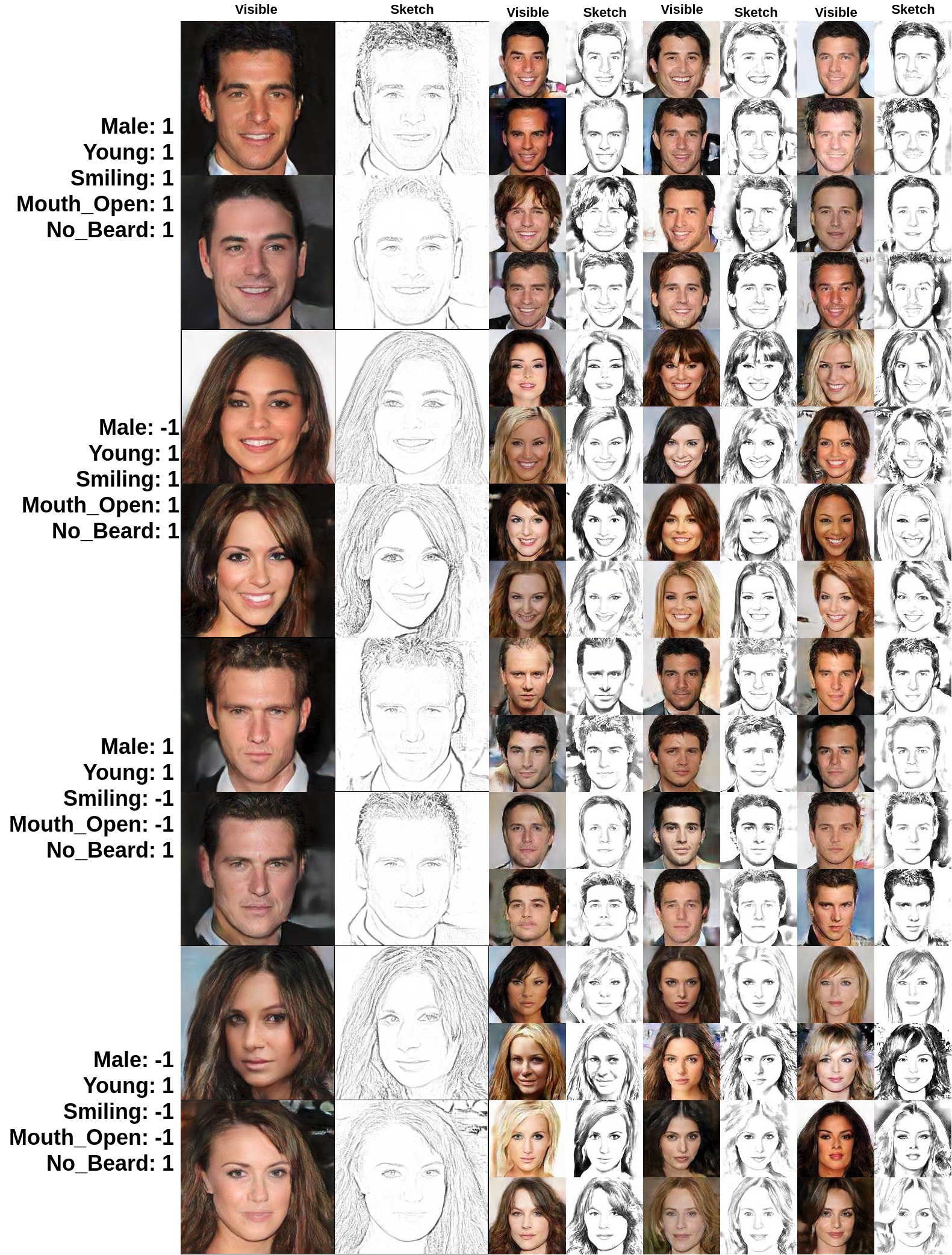}
		\caption{Additional 256x256 and 128x128 multimodal images generated using the CELEBA-HQ dataset.}
		\label{fig:additional_CelebA}
	\end{figure*}

\begin{figure*}[htp!]
		\centering
		\includegraphics[width=0.9\linewidth]{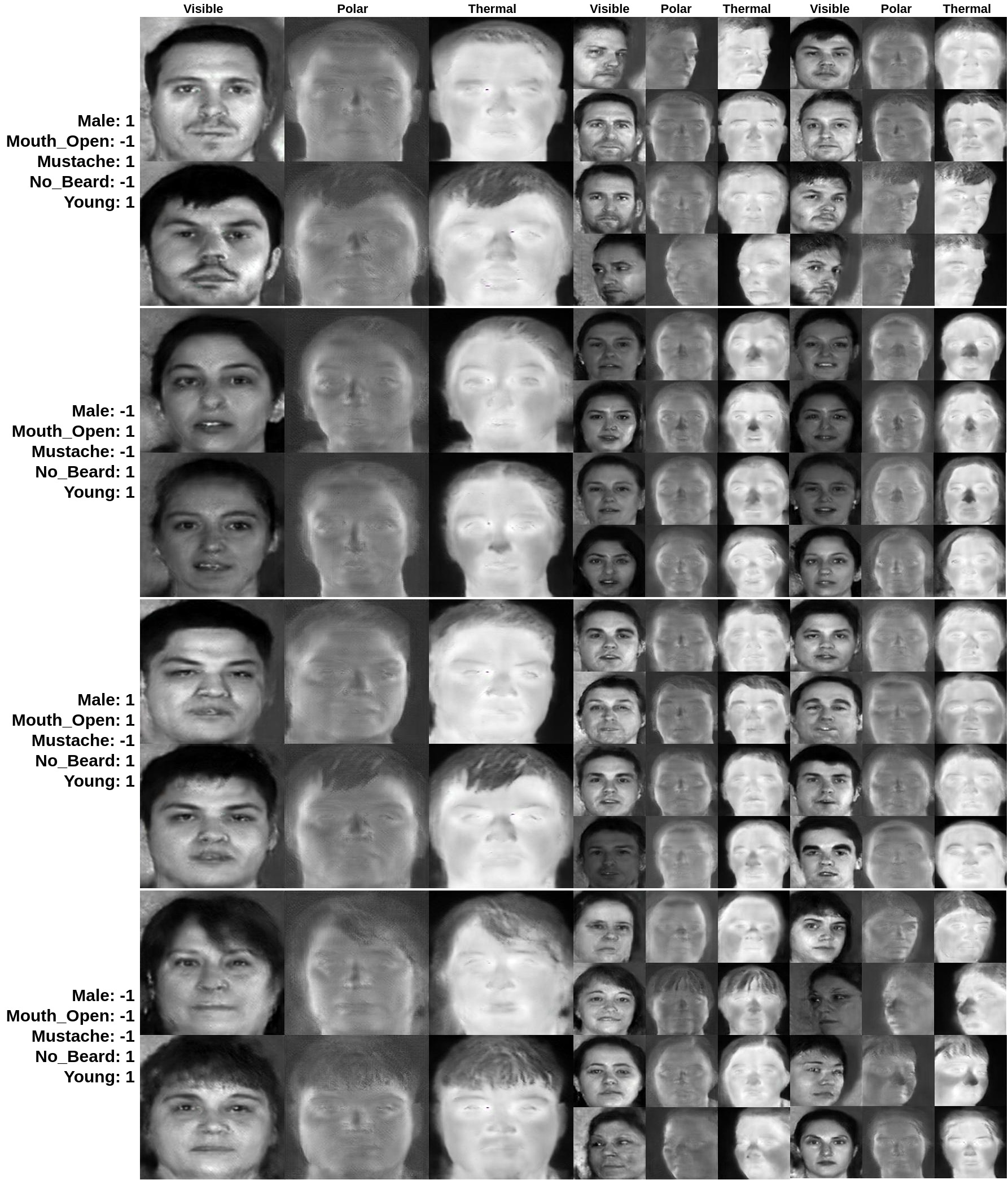}
		\caption{Additional 256x256 and 128x128 multimodal images generated using the CASIA-NIR-VIS 2.0 dataset.}
		\label{fig:additional_CASIA}
	\end{figure*}

	Additional multimodal synthesized samples corresponding to the proposed method from the ARL Multimodal Face dataset, CelebA-HQ dataset and CASIA-NIR-VIS 2.0 dataset can be found in Figure~\ref{fig:additional_Casia}, \ref{fig:additional_CelebA}, and \ref{fig:additional_CASIA}, respectively.

	\begin{figure*}
		\centering
		\includegraphics[width=0.95\linewidth]{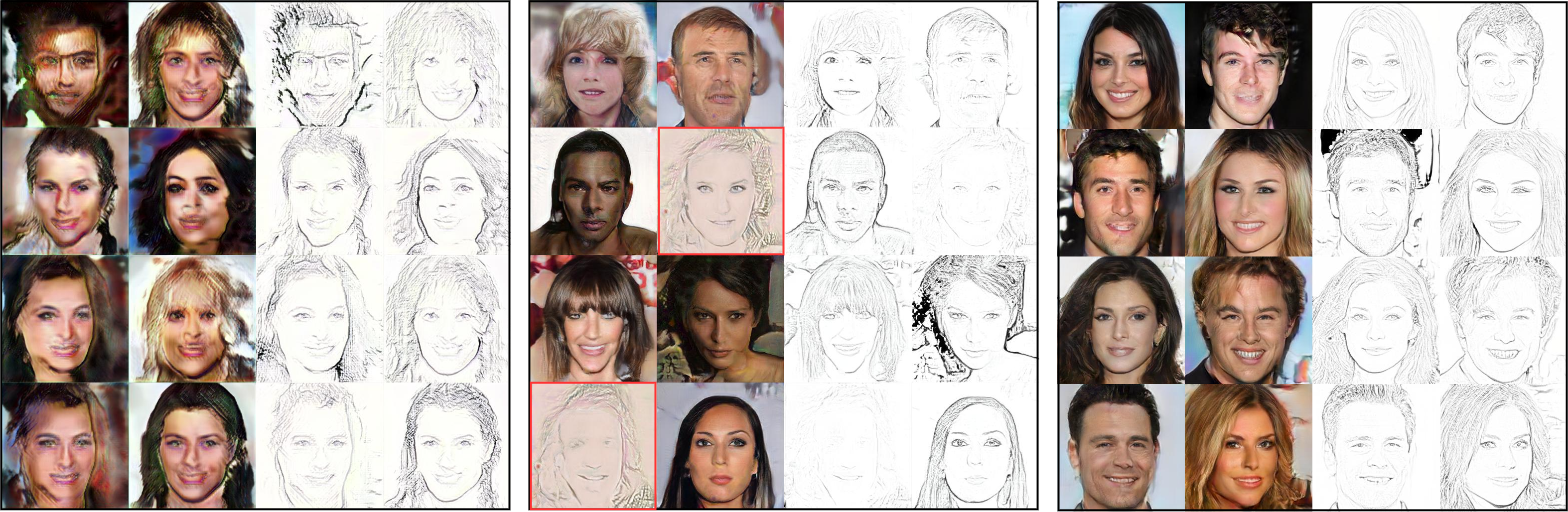} \\
		\raggedright \hskip 50pt (a) Batch Norm  \hskip 100pt (b) Instance Norm \hskip 90pt (c) Feature Norm 
		\caption{Comparison of using different normalization methods. The FID scores for visible/sketch modalities in different normalization methods are: (a) Batch Norm:118.69/96.01; (b) Instance Norm:44.82/28.01; (c) Feature Norm: 13.30/17.75. The modality implications are highlighted with \textcolor{red}{red boxes}}
		\label{fig:multimodalnormalizationcomparison}
	\end{figure*}

	\noindent {\bf{Attribute Accuracy:}}  In order to check whether the generated images preserve the attributes that were used to synthesize images, we use mean square error (MSE) between the predicted attribute scores corresponding to the  reference  image  and  the  synthesized  image. We fine-tune ResNet50 \cite{he2016deep} on the attribute label by replacing the final layer of the classifier. The estimated attributes from the re-trained ResNet50 model are used for evaluation. When conducting this experiment, 5000 ground-truth images with corresponding attributes are randomly selected. The averaged score and standard derivation based on 5 splits are shown in Table~\ref{tab: attribute accuracy}. As can be seen from this  table, the proposed Att2MFace is able to preserve the attributes on the synthesized images better than the other baselines.  Note that since extracting facial attributes from sketch or thermal modalities is difficult, we only estimate attributes from the synthesized visible images. 
	
	\noindent {\bf{Normalization Comparison:}}    In order to demonstrate the effectiveness of feature normalization in Eq.~\eqref{eq:feature_equalization}, we compare it with two common normalization methods:  batch normalization and  instance normalization. For fair comparison, we replace the feature equalization with these two normalizations and keep the other network structure the same. We show the performances visually in Figure ~\ref{fig:multimodalnormalizationcomparison}. As can be seen from this figure instance normalization leads to modality implication (highlighted in red boxes).  In addition, the FID scores are indicated in the caption to show the superiority of feature equalization quantitatively.    
	\DIFaddbegin

	\DIFaddend

	\begin{table}[htp!]
		\centering
		\caption{Attribute accuracy based on the MSE metric with mean $\pm$ std. }
		\label{tab: attribute accuracy}
		\resizebox{1.0\columnwidth}{!}{%
			\begin{tabular}{|c|c|c|c|}
				\hline
				Methods & \multicolumn{1}{c|}{ARL Multimodal Face Database}          & \multicolumn{1}{c|}{CelebA Database} & 		 \multicolumn{1}{c|}{CASIA NIR-VIS 2.0} \\ \hline
				CoGAN \cite{liu2016coupled} & 0.0583 $\pm$ 0.0181 & 0.0671 $\pm$ 0.0152 & 0.0603 $\pm$ 0.0135 \\ \hline
				
				RegCGAN \cite{mao2018unpaired} & 0.0594 $\pm$ 0.0146 & 0.0583 $\pm$ 0.0132 & 0.0574 $\pm$ 0.0177 \\ \hline
				
				StarGANv2$^{\star}$ \cite{choi2020stargan} & 0.0547 $\pm$ 0.0142 & 0.0563 $\pm$ 0.0094 & 0.0553 $\pm$ 0.0136 \\ \hline
				
				Att2MFace (ours) & \textbf{0.0516 $\pm$ 0.0158} & \textbf{0.0489$\pm$ 0.0133} & \textbf{0.0508 $\pm$ 0.0148} \\ \hline
				
			\end{tabular}
		}
	\end{table}
	
	\begin{figure}[t]
		\centering
		\includegraphics[width=0.95\linewidth]{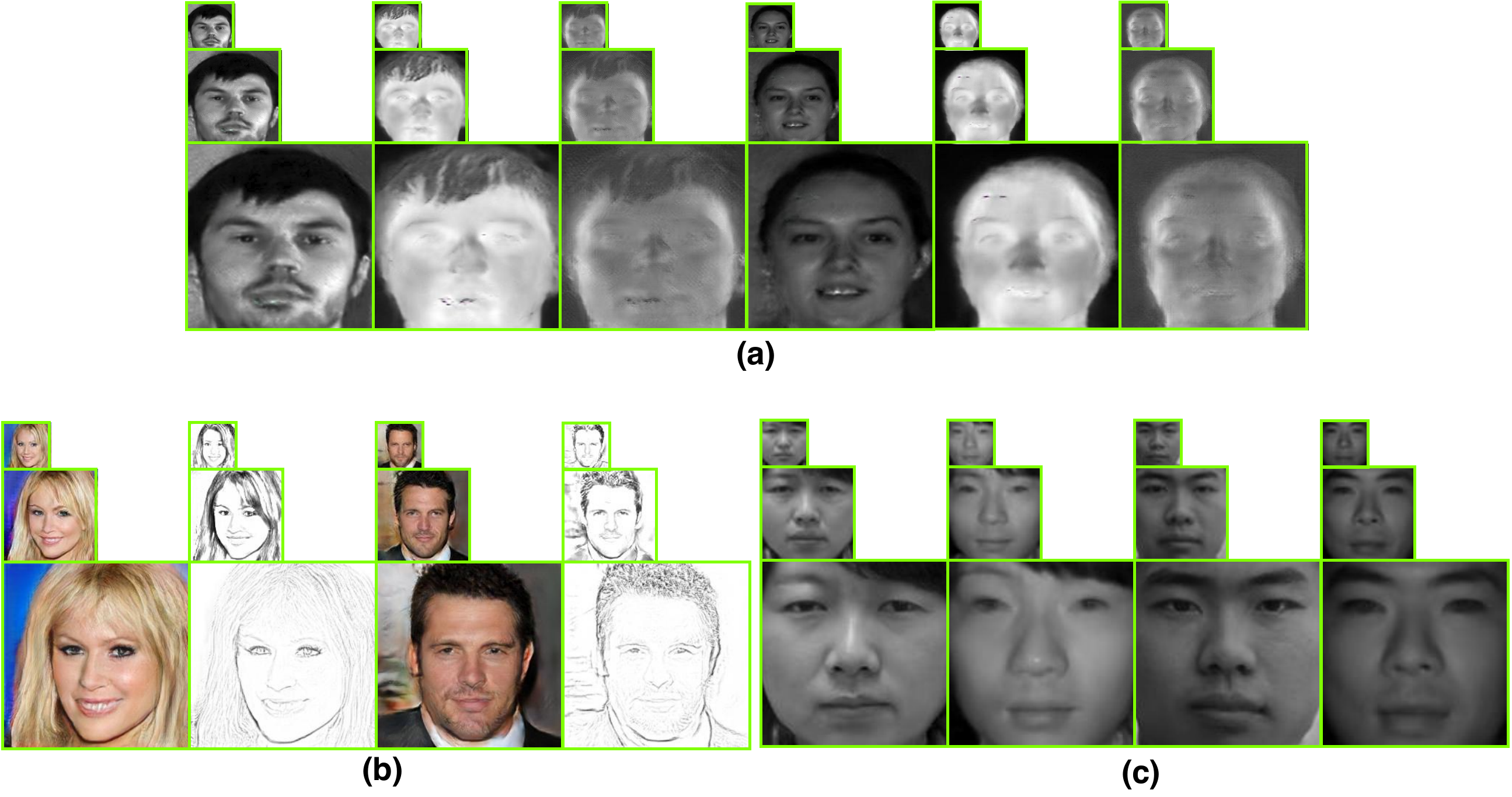}
		\vskip -8pt \caption{Synthesized multimodal images during progressive-growth training at different resolutions. }
		\label{fig:progressivegrowth2}
	\end{figure}
	
	\noindent {\bf{Progressive Learning:}}  Fig.~\ref{fig:progressivegrowth2} shows sample outputs during progressive training of our network.  As can be seen from this figure, starting from low-resolution, our method progressively increases the resolution of multimodal images. This incremental learning framework allows the network to discover the overall structure of the face first and then adds finer scale detail at larger scales.  Hence, it avoids having to learn multimodal images at all scales simultaneously.  This process also stabilizes the training process.

	\subsection{ Face Synthesis via Manipulating }
\begin{figure}[t]
		\centering
		\includegraphics[width=0.95\linewidth]{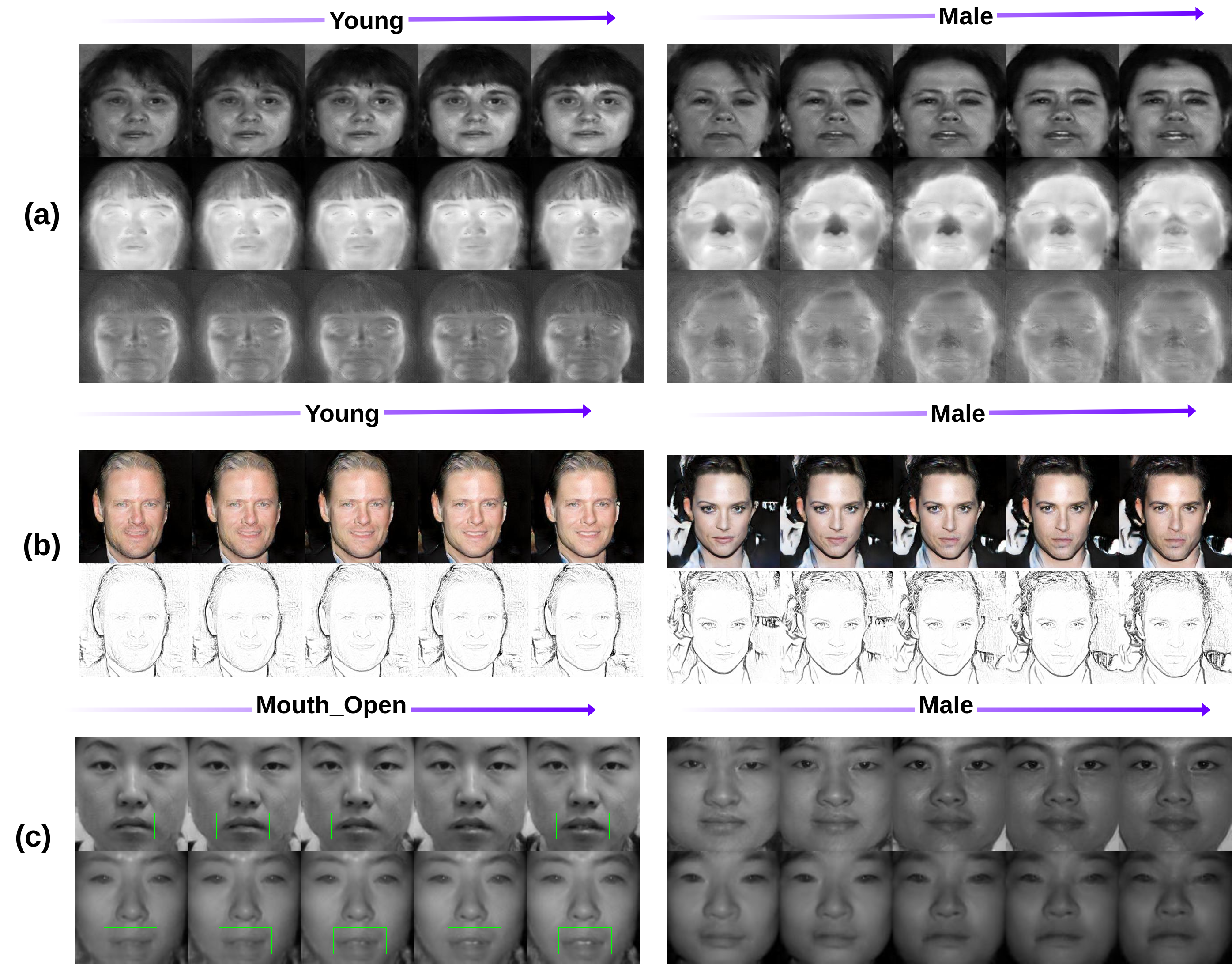}
		\caption{Progressive synthesis of multimodal face images when attributes are changed while the noise vector is kept fixed. (a)Old to young and female to male synthesis on the ARL dataset.  (b) Old to young and female to male synthesis on the CelebA-HQ dataset. (c) Mouth closed to open and female to male synthesis on the CASIA-NIR-VIS dataset.}
		\label{fig:manipulateattribute}
	\end{figure}

\begin{figure}[htp!]
		\centering
		\includegraphics[width=0.9\linewidth]{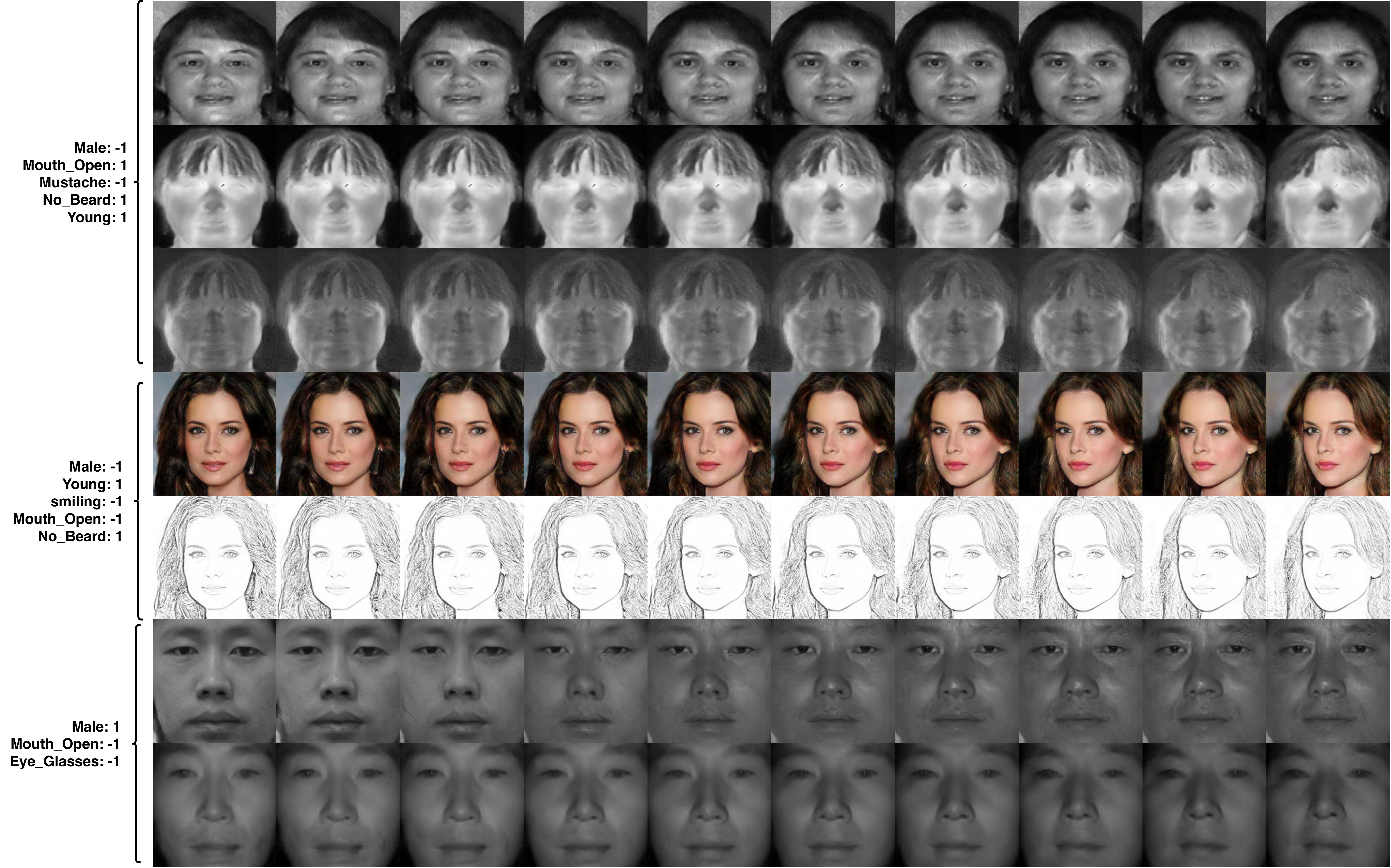}
		\caption{Synthesis of multimodal images  via interpolation between two noise codes with fixed visual attribute (listed on the left-side).  Note that  the identity and facial shape change as we vary the noise vector but the attributes are preserved  on the synthesized images}
		\label{fig:manipulatenoise}
	\end{figure}

	To understand whether the model learns to generate a diverse set of images or just memorizes data, two experiments are conducted.  In the first experiment, we manipulate the attribute code while keeping the noise vector fixed.  This experiment shows the  image  synthesis  capability  of our network by manipulating the input attribute.  In the second experiment, we fix the attribute code and change the noise vector.  This experiment shows whether our model can learn a smooth mapping from noise to the image space.\\
	\noindent 	\textbf{Attribute Manipulation:}  Given an attribute code $\mathbf{y}_{a}$, another attribute code $\mathbf{y}^{\star}_{a}$ is obtained by flipping a particular value of $\mathbf{y}_{a}$ (i.e. Male: -1 $\rightarrow$ Male: 1). The attribute code is interpolated as $\mathbf{y}_{a} = \beta * \mathbf{y}_{a} + (1-\beta) * \mathbf{y}^{}_{a}$ with $\beta\in [0,1]$.  The manipulated attribute code is then used to synthesize multimodal images by keeping the noise vector fixed.  Results corresponding to this experiment are shown in Fig.~\ref{fig:manipulateattribute}.  As can be seen from this figure, when higher weights are given to a certain  attribute,  the  corresponding  appearance changes.  For instance, we are able to synthesize  multimodal   images  with  a different  gender  by  changing  the  weights  corresponding  to the  gender  attribute  as  shown  in the first three rows of Fig.~\ref{fig:manipulateattribute}. Each  column  shows the progression of gender change as the attribute weights are manipulated from -1 to 1.   Similarly, the  synthesis  results when young and mouth open attributes are changed are also shown in Fig.~\ref{fig:manipulateattribute}.  It is worth noting that when  the attribute weights other than the gender attribute are changed, the identity of the person does not change much.\\
	\noindent\textbf{Noise Manipulation:} Similar to the above experiment, we gradually synthesize images from the interpolated latent vectors between the two noise codes while keeping the  attribute code frozen.  Results corresponding to this experiment are shown in Fig.~\ref{fig:manipulatenoise}.  As can be seen from this figure, as we change the noise vector, attributes stay the same while the identity changes.  Our model is able to learn the mapping from the latent noise space instead of just memorizing the data. 
	
	\section{Conclusions}
	We proposed a novel network architecture, called Att2MFace, for multimodal face generation from visual attributes. Att2MFace consists of a single generator that can simultaneously generate multimodal face images from visual attributes.  Furthermore, we take advance of the progressive training strategy to synthesize consistant multimodal face images. Qualitative and quantitative comparison with other state-of-the-art methods on three datasets demonstrate the superiority of the proposed method. Various experiments showed that the proposed method  is  able  to  generate  high-quality multimodal images  and  achieves significant improvements over the state-of-the-art methods.
	


	{\small
		\bibliographystyle{ieee}
		\bibliography{multimodal.bib}
	}

\end{document}